\documentclass[10pt,twocolumn,letterpaper]{article}

\usepackage{iccv}

\makeatletter
\@namedef{ver@everyshi.sty}{}
\makeatother

\usepackage{times}
\usepackage{epsfig}
\usepackage{graphicx}
\usepackage{amsmath}
\usepackage{amssymb}
\usepackage{tabularx}
\usepackage{multirow}
\usepackage[numbers]{natbib}
\usepackage{todonotes}

\usepackage[pagebackref=true,breaklinks=true,letterpaper=true,colorlinks,bookmarks=false]{hyperref}

\iccvfinalcopy 

\def\iccvPaperID{8543} 

\newcolumntype{Y}{>{\centering\arraybackslash}X}

\newcounter{alphasect}
\def\alphainsection{0}

\let\oldsection=\section
\def\section{%
  \ifnum\alphainsection=1%
    \addtocounter{alphasect}{1}
  \fi%
\oldsection}%

\renewcommand\thesection{%
  \ifnum\alphainsection=1%
    \Alph{alphasect}%
  \else
    \arabic{section}%
  \fi%
}%

\newenvironment{alphasection}{%
  \ifnum\alphainsection=1%
    \errhelp={Let other blocks end at the beginning of the next block.}
    \errmessage{Nested Alpha section not allowed}
  \fi%
  \setcounter{alphasect}{0}
  \def\alphainsection{1}
}{%
  \setcounter{alphasect}{0}
  \def\alphainsection{0}
}%

\begin{document}

\title{A Semantic Segmentation Network for Urban-Scale Building Footprint Extraction Using RGB Satellite Imagery}

\author{Aatif Jiwani\\
UC Berkeley\\
Lawrence Berkeley Lab\\
{\tt\small aatifjiwani@lbl.gov}
\and
Shubhrakanti Ganguly\\
UC Berkeley\\
Lawrence Berkeley Lab\\
{\tt\small shubhra@lbl.gov}
\and
Chao Ding \thanks{Corresponding Author: Please send correspondance to \texttt{chaoding@lbl.gov} } \\
Lawrence Berkeley Lab\\
{\tt\small chaoding@lbl.gov}
\and
Nan Zhou\\
Lawrence Berkeley Lab\\
{\tt\small nzhou@lbl.gov}
\and
David Chan\\
UC Berkeley\\
{\tt\small davidchan@berkeley.edu}
}

\maketitle

\begin{abstract}
Urban areas consume over two-thirds of the world’s energy and account for more than 70\% of global $CO_2$ emissions. As stated in IPCC's Global Warming of 1.5 ºC report, achieving carbon neutrality by 2050 requires a clear understanding of urban geometry. 
High-quality building footprint generation from satellite images can accelerate this predictive process and empower municipal decision-making at scale. However, previous Deep Learning-based approaches face consequential issues such as scale invariance and defective footprints, partly due to ever-present class-wise imbalance. Additionally, most approaches require supplemental data such as point cloud data, building height information, and multi-band imagery - which has limited availability and are tedious to produce. In this paper, we propose a modified DeeplabV3+ module with a Dilated Res-Net backbone to generate masks of building footprints from three-channel RGB satellite imagery only. Furthermore, we introduce an F-Beta measure in our objective function to help the model account for skewed class distributions and prevent false-positive footprints. In addition to F-Beta, we incorporate an exponentially weighted boundary loss and use a cross-dataset training strategy to further increase the quality of predictions. As a result, we achieve state-of-the-art performances across three public benchmarks and demonstrate that our RGB-only method produces higher quality visual results and is agnostic to the scale, resolution, and urban density of satellite imagery.\footnote[1]{Code is publicly available at https://github.com/aatifjiwani/rgb-footprint-extract/}.
\end{abstract}


\section{Introduction}
\begin{figure*}[h!]
\begin{center}
\includegraphics[width=\textwidth]{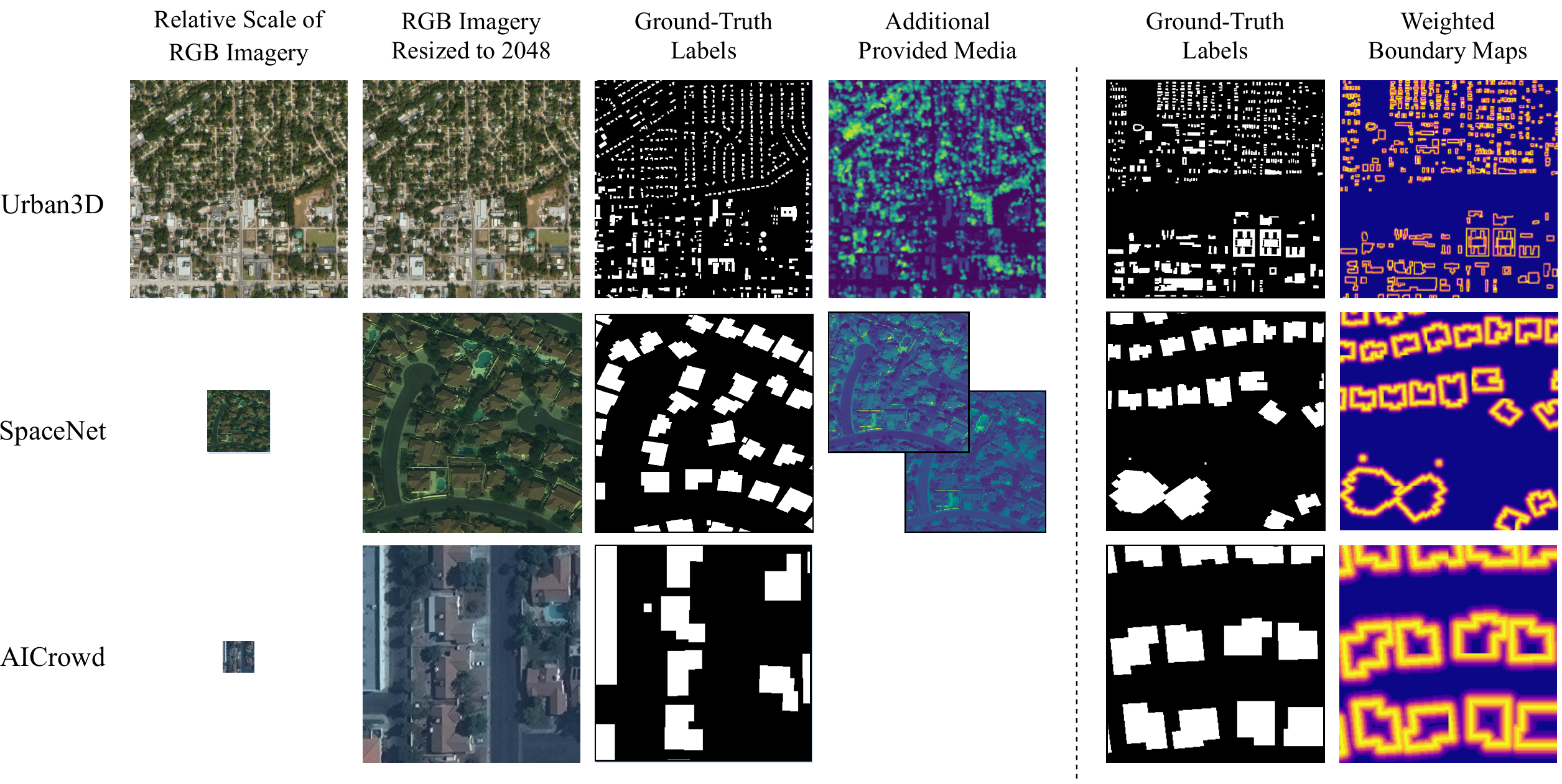}
\end{center}
  \caption{\textbf{Left:} Visualizations of samples from each dataset used in this paper (see Section \ref{section:datasets} for details). \textbf{Right:} Visualizations of weighted boundary maps ($w_0 = 10$, $\sigma = 7.5$). }
\label{fig:dataset}
\label{fig:expweightviz}
\end{figure*}

Urban centers consume over two-thirds of the world’s energy and account for more than 70 percent of global $CO_2$ emissions. Achieving carbon neutrality by 2050, as stated in IPCC's Global Warming of 1.5°C report \cite{ipcc15}, will require a good understanding of the urban geometry at speed and scale \cite{climatechange2015}. Automatic building footprint extraction from satellite imagery is currently being pursued to support many urban science applications, such as urban planning, energy efficiency, micro-climate modeling and emergency response \cite{chaonan, chaolam, chen2018benchmark}. Furthermore, in high-density urban regions where buildings are often close to one another, an accurate distinction of adjacent buildings is required to provide any meaningful support to architects and urban planners. However, although building footprint extraction has received attention from the Deep Learning community \cite{crowdAI, spacenet, urban3d, ssgisdata, cvprspacenetunetensemble, cvprspacenetmrcnn, audebert}, approaches based on convolutional neural networks (CNN) continue to face prominent issues such as scale invariance and defective footprints. Building density and resolution can vary significantly among satellite imagery, as can be seen in Figure \ref{fig:dataset}, presenting class imbalances that make it difficult for a CNN to be robust on unseen data. Consequently, CNN-based approaches commonly suffer from sparse false positives and adjacent footprints being predicted as a single entity, drastically reducing the level of insight urban planners can glean from results.\par

The Deep Learning community has made many approaches that attempt to resolve conjoined and extraneous footprints, however, most approaches require supplemental data such as point clouds, building height information, and multi-band imagery - all of which are too expensive to produce or unattainable for most cities worldwide. Contrary to these methods, we propose a novel method that relies on easily attainable and globally available RGB satellite images. In the following sections, we present a modified DeepLabV3+ architecture and unique two-part training objective, and demonstrate that our method not only performs better in numerical performance, but also produces cleaner footprint predictions and precisely separates buildings in densely populated areas.


\section{Related Work}
\subsection{Deep Learning and Building Segmentation}
Building segmentation in remote sensing has long been established as an important task in urban planning, inviting robust non-Deep Learning approaches such as LIDAR reconstruction \cite{lidarrecon}. However, with processing power becoming progressively cheaper and more accessible, Deep Learning approaches for this task are becoming increasingly prominent and successful. Li \etal \cite{ssgisdata} tackle urban building segmentation with a U-Net ensemble and attempt to resolve the scale invariance problem by splitting their ensemble into two sets: one receives a downsampled copy, and the other receives a stack of 9 equal sections from the original image. Delassus \etal \cite{cvprspacenetunetensemble} also use a U-Net ensemble, but move in a different direction and attempt to separate footprints by employing the DICE Coefficient \cite{sorenson, dice}. In our work, we demonstrate that replacing the DICE Coefficient with the F-Beta measure has numerical and visual benefits. Marmanis \etal \cite{marmanis} employ a simple multi-layer perceptron, and attempt to separate conjoined footprints by framing the task as a three-class classification problem. Similarly, Yang \etal \cite{yang2018} employ a SegNet \cite{segnet} and developed a signed-distance function to encode relative distances between buildings within ground-truth masks. Using the distance function, they transform the binary masks into masks with 128 classes wherein pixels that are closer to buildings have a higher class than pixels that are generally background. Inspired by these approaches, in our work we introduce an exponentially weighted boundary loss to our objective that heavily penalizes incorrect boundary predictions to achieve better separation of buildings in close proximity. Instead of manipulating the objective, Zorzi \etal \cite{zorzi} propose a pipeline that performs regularization and polygonization on segmentation masks to produce clear-cut footprints. Unlike any of the previous approaches, \cite{zorzi} employ generative adversarial networks with the intention of cleaning up segmented masks before polygonization \cite{GANpaper}. However, as Zorzi \etal note, this method often hinders performance when buildings are occluded by the presence of greenery, creating skewed footprints that do not properly overlap with the ground-truth.

\subsection{Loss Functions for Data Imbalance}
Class-wise imbalance in the data is a tough problem that presents itself in multifarious tasks, but we crucially highlight its presence in object detection and image segmentation. In object detection, there is an inherent imbalance as foreground objects are often just a small fraction of the full image. Lin \etal \cite{focal} address the difficulty of training object detectors with the Focal loss. Commonly applied in object detection tasks, the Focal loss builds upon binary Cross-Entropy by adding a tunable factor that effectively puts more emphasis on penalizing misclassifications by shifting focus away from the common class. However, the Focal loss presupposes the common class is unimportant, which does not translate well to building segmentation as we observed copious false positives footprints. In binary image segmentation, the common problem of class imbalance poses a downstream problem for either precision or recall. Salehi \etal \cite{tversky} developed the Tversky index to shift the network's focus on resolving false negatives. The Tversky index, based upon the DICE coefficient \cite{sorenson, dice}, is an ornate function of the predicted and ground-truth masks with tunable weights $\alpha$ and $\beta$ that shift emphasis towards precision and recall respectively. Compared to \cite{focal} and \cite{tversky}, we introduce the F-Beta measure which requires only a single parameter $\beta$ to analogously shift focus towards precision or recall while not losing sight of the common class. Finally, Kervadec \etal \cite{kervadec2019boundary} proposed a distance-function-based boundary loss to resolve imbalances in brain scans. While \cite{kervadec2019boundary} have shown they improved the visual quality of results, the issue of separating distinct entities remains extant. In this work, we introduce a boundary loss aimed to ameliorate the separation of distinct ground-truth footprints.

\section{Approach}
\label{section:approach}

We propose a network architecture and training method closely modeled on \cite{ssgisdata} and \cite{cvprspacenetunetensemble}, however, with several major alterations to account for the strict use of RGB imagery:
\begin{enumerate}
    \itemsep0em
    \item We replace the U-Net encoder-decoder model with a DeepLabV3+ module \cite{deeplabv3}.
    \item We swap the Aligned Xception model in the DeepLabV3+ with a Dilated ResNet \cite{dilatedresnet}.
    \item We generalize the DICE loss to an F-Beta Measure and add an exponentially weighted boundary loss to create a unique two-part training objective.
    \item We perform cross-dataset training to optimize performance on multiple datasets.
\end{enumerate} 

\begin{figure*}
\begin{center}
\includegraphics[width=\textwidth]{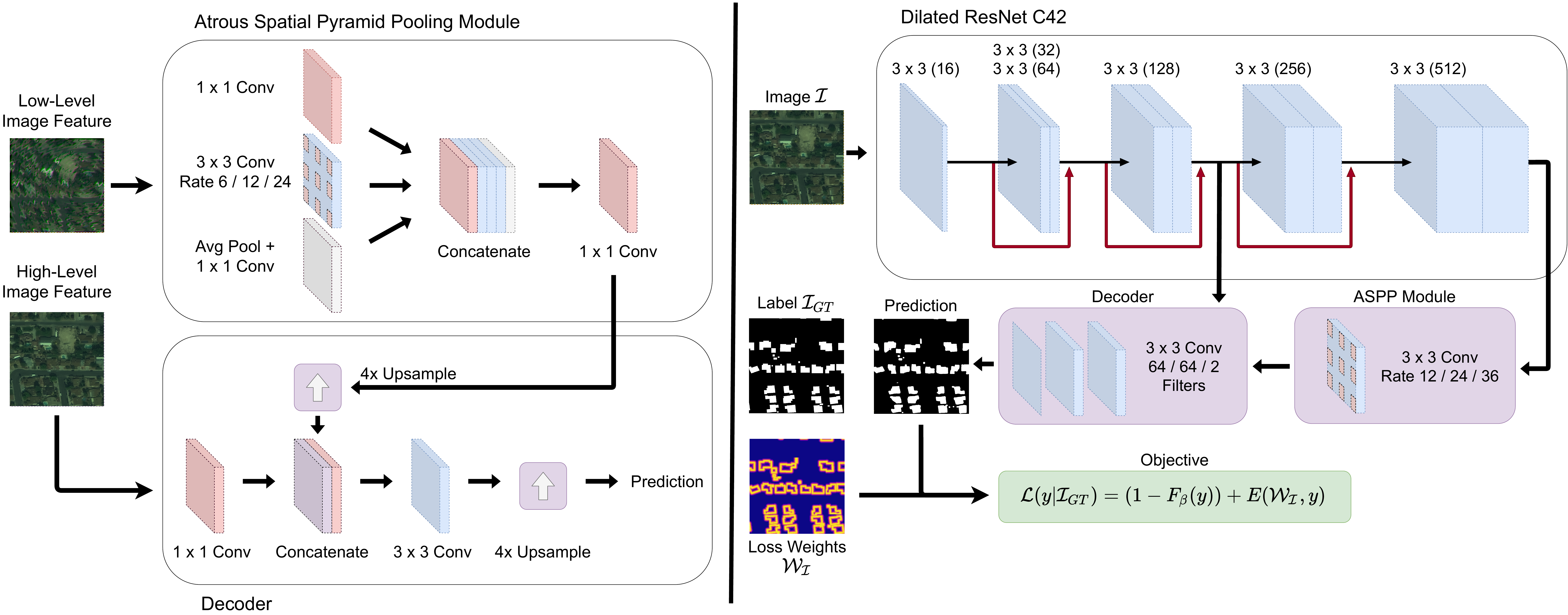}
\end{center}
  \caption{\textbf{Left: } DeepLabV3+ Module as depicted by \cite{deeplabv3}. \textbf{Right: } Training pipeline depicting our modified semantic segmentation network.}
\label{fig:architecture}
\end{figure*}

We depict our overall training pipeline in Figure \ref{fig:architecture} and compare it side-by-side against the default DeepLabV3+ module. 

\subsection{Modified DeepLabV3+ Network}
\label{section:deeplab}

In our work, we found that for RGB data, as opposed to wideband satellite imagery, the DeepLabV3+ architecture outperformed the U-Net based method. We hypothesize that this is due to DeepLabV3+'s increased ability to model local and global relationships between parts of the visual data, which stems from two key features in the architecture. First, the spatial pyramid pooling helps encode multi-scale contextual information in the model. Second, skip connections in the encoder, with low-level features from the backbone, help refine the prediction by allowing details (edges and corners of buildings) to permeate the final layers.

We also found that by further replacing the Xception Module in the DeepLabV3+ architecture with dilated convolutions, we can reduce the loss of spatial context information by introducing a larger receptive field (RF) for each kernel image. This added scale invariance is clear in our approach (See Figure \ref{fig:visualpredictions}). Dilated convolutions consist of kernels with “holes” in them to achieve better performance in downstream tasks, allowing for a larger RF in earlier layers without drastically increasing the number of parameters. Downsampling and pooling in the encoder can reduce spatial context in deeper layers, which the larger RF mitigates and allows us to track less prominent features (ex. smaller or narrow buildings). 

\subsection{Objective Function}
\label{section:objective}


\begin{equation} \label{eq:old}
    \mathcal{L}(y, \hat{y}) = \mathcal{L}_{BCE}(y, \hat{y}) + (1 - D(y, \hat{y}))
\end{equation}

\noindent Methods for semantic segmentation commonly optimize Equation (\ref{eq:old}), where $D$ is the DICE coefficient (a generalization of the F1 Score)\cite{surveyloss}. Works that utilize Eqn. (\ref{eq:old}) achieve relatively high numerical accuracy on the satellite segmentation datasets, but encounter visual issues such as combined footprints and sparse false building predictions (See Figure \ref{fig:visualpredictions} "BCE + 1" column). This hinders the analysis of urban areas because we may not obtain an accurate qualitative prediction of the building masks required for downstream energy estimation. We speculate these visual inaccuracies are due to two major causes. First, since the resolution of the data is not always high, the building-to-background distinction can be blurry, making it hard to differentiate tightly packed buildings. Second, false-positive building clusters are abundantly predicted in footprint masks, which stems from the distribution of building-to-background pixels being uneven (see Figure \ref{fig:dataset}), with far more background than building. We address both of these drawbacks with a modified two-part loss formulated as follows:

\begin{equation}
\label{eq:ourobjective}
    \mathcal{L}(y, \hat{y}) = \mathcal{L}_{EWC}(y, \hat{y}) + (1 - F_{\beta}(y, \hat{y})) 
\end{equation}

\noindent{where}

\begin{equation}
\label{eq:ourFbeta}
    F_{\beta}(y, \hat{y}) = \frac{(1 + \beta^2)(y * \hat{y})}{\beta^2(y + \hat{y})}
\end{equation}

\noindent{and}

\begin{equation}
\label{eq:wcelossterm}
    \mathcal{L}_{WCE}(y, \hat{y}) = \sum_{(i,j)} \mathcal{W}_{ij}(y)*log(p_{y_{ij}}(\hat{y}_{ij}))
\end{equation}

\noindent\textbf{F-Beta Measure as a Loss Function:} Due to imbalances in the training data, high-parameter models such as the DeepLabV3+ tend to lean toward predicting buildings rather than the more common background classes.
To compensate for the erroneous false positives, we introduce the F-Beta Measure to replace the DICE coefficient in our two-part objective. Rooted from the general F1 Score, we define the F-Beta Measure in Equation (\ref{eq:ourFbeta}) as a configurable loss function with a hyper-parameter $\beta$. However, although Eqn. (\ref{eq:ourFbeta}) is a weighting of \cite{dice}, this weighting helps the F-Beta Measure serve as a crucial \textbf{generalization} of the DICE coefficient. Note that when we set $\beta = 1$, the F-Beta Measure precisely becomes the DICE coefficient. Here, we argue that this simple yet unique formulation of the F-Beta Measure allows us to \textbf{tune} our objective toward prioritizing either false positives or false negatives. We will show that by setting $\beta > 1 $, we place a higher emphasis on preventing false negatives, thus increasing recall. Likewise, if we set $\beta < 1$, we force the network to focus on improving precision, consequently resolving false positives and thus reducing the number of misplaced building clusters. \\

\noindent\textbf{Exponential Weighted Boundary Loss (EWC):} Similar to \cite{marmanis}, which introduces a third building boundary class, we add a weighted loss function \cite{unet} that heavily penalizes wrong predictions close to and on building edges. Our exponential boundary loss generates a weight map $w(\boldsymbol{x})$ that forces the model to separate neighboring objects into distinct entities:

\begin{equation}
    w(\boldsymbol{x}) = w_c(\boldsymbol{x}) + w_0 \cdot exp\bigg(-\frac{(d_1(\boldsymbol{x}) + d_2(\boldsymbol{x}))^2}{2\sigma^2} \bigg) 
\label{eq:unetweight}
\end{equation}

\noindent where $\boldsymbol{x}$ is a ground-truth mask, $w_c(\boldsymbol{x})$ is a separate weight map used to resolve class imbalance, and $w_0$ and $\sigma$ are constants that help tune the absolute loss value and relative decay respectively. Using this mask not only helps us distinguish adjacent buildings, but also increases the overall loss in high-density areas (ex. Figure \ref{fig:expweightviz} top-right) where models traditionally exhibit a greater level of uncertainty (see Section \ref{section:results}). 

The evaluation of Equation (\ref{eq:unetweight}) empirically produces values between $0$ and $w_0$ per-pixel. Although we want to heavily penalize incorrect boundary predictions, we still want to properly penalize \textit{all} incorrect background predictions. Therefore, we build upon Equation (\ref{eq:unetweight}) to produce a new weight map $\mathcal{W}(\boldsymbol{y})$

\begin{equation}
    \mathcal{W}(\boldsymbol{y}) = exp(w(\boldsymbol{y}))^p
\label{eq:unetweightnew}
\end{equation}

\noindent that produces values between $1$ and $e^p$, where $p$ is an additional hyper-parameter to further prevent incorrect boundary predictions without affecting how we consider less important background predictions. Figure \ref{fig:expweightviz} visualizes these new weight maps on the three standard datasets. We incorporate $\mathcal{W}(\boldsymbol{y})$ as part of the objective in Equations (\ref{eq:ourobjective}) and (\ref{eq:wcelossterm}) and claim it allows the network to separate previously conjoined predictions. \\

\noindent\textbf{Cross-Dataset Training Strategy:} As Section \ref{section:datasets} will quantify, there is a stark imbalance in the number of samples across the datasets. Consequently, we saw quick over-fitting and lack of quality results in Urban3D. To account for this lack in overall performance, we adopt a cross-dataset training strategy \cite{zhang2017transfer}. We first train on a synthesized set that combines augmented samples from the datasets, and then further fine-tune on the individual dataset. We claim and subsequently prove that by training on similar satellite images before fine-tuning on the individual scale, we achieve better numerical performance and more distinct visual results. 

\section{Experiments}

\subsection{Datasets}
\label{section:datasets}
\noindent We evaluate our method on three datasets. Each dataset is split into an (80-20)-20 partition, where 20\% is reserved for the final test set, and the remaining 80\% is further split into (80-20) for training and validation sets, respectively.

\noindent\textbf{SpaceNet:} The SpaceNet.ai Building Detection Dataset (SpaceNet) \cite{spacenet} contains 700,000 building footprints in five cities scattered across the globe. In this study, we evaluate using the Vegas region which contains over 3,800 samples of 200m x 200m areas. Each input sample comes with an associated 650 x 650 pixel RGB satellite image, a high-resolution panchromatic image, and a low-resolution multi-spectral image. For our purposes, we only use the RGB satellite image. The samples in their original form come in a TIFF raster graphics format and the labels in geographical coordinates, so we use GDAL to pre-process all samples by extracting RGB rasterizations and converting the coordinates into grayscale masks \cite{osgeo}.  \\
\noindent\textbf{AICrowd:} The AICrowd Mapping Challenge dataset (AICrowd) \cite{crowdAI} contains 340,000 total samples as 300 x 300 pixel RGB images. RGB image samples are provided in JPEG format, and annotations are provided in MS-COCO format for which we use the corresponding API to extract masks \cite{mscoco}.  \\
\noindent\textbf{Urban3D:} TopCoder’s Urban3D dataset (Urban3D) \cite{urban3d} contains 236 samples of 2048 x 2048 pixel images and labels, both in the TIFF compressed raster format. We use similar pre-processing steps as SpaceNet to extract RGB images and grayscale masks. As a similar note to SpaceNet, each RGB sample in this dataset is accompanied by its Depth Surface Model and Digital Terrain Model, which provides high-resolution building height information. As mentioned with SpaceNet, we use the RGB image for our purposes only. 

\begin{table}
\footnotesize
\begin{tabularx}{\linewidth}{|Y|c|p{20pt} p{20pt}p{20pt}p{20pt}|} 
\hline
\empty & Model & AP & AR & F-1 & mIOU \\
\hline\hline
\multirow{4}{4em}{Urban3D} & \textbf{Dilated-RN$_{D}$} & 82.0 & \textbf{78.5} & \textbf{79.6} & \textbf{81.0} \\ 
& {Res-Net$_{D}$} & 81.5 & 77.3 & 78.6 & 80.1\\ 
& {U-Net} & \textbf{83.1} & 76.2 & 78.6 & 80.5\\
& {FCN} & 82.7 & 75.5 & 78.2 & 80.0 \\ 
\hline
\multirow{4}{4em}{SpaceNet} & \textbf{Dilated-RN$_{D}$} & 90.0 & \textbf{91.1} & \textbf{90.6} & \textbf{89.1} \\ 
& {Res-Net$_{D}$} & 90.2 & 89.9 & 90.1 & 89.0 \\ 
& {U-Net} & \textbf{90.6} & 89.1 & 89.7 & 88.2 \\
& {FCN} & 90.1 & 88.9 & 89.5 & 88.4 \\ 
\hline
\multirow{4}{4em}{AICrowd} & 
\textbf{Dilated-RN$_{D}$} & { \scriptsize \textbf{90.7$_{.02}$}} & {\scriptsize \textbf{90.0$_{.01}$}} & {\scriptsize \textbf{90.4$_{.04}$}} & {\scriptsize \textbf{88.6$_{.00}$}} \\ 

& {Res-Net$_{D}$} & {\scriptsize 90.2$_{.10}$} & {\scriptsize 89.0$_{.02}$} & {\scriptsize 89.7$_{.07}$} & {\scriptsize 87.9$_{.00}$} \\ 

& {U-Net} & {\scriptsize 88.4$_{.16}$} & {\scriptsize 86.3$_{.06}$} & {\scriptsize 87.0$_{.08}$} & {\scriptsize 85.3$_{.00}$}  \\

& {FCN} & {\scriptsize 89.6$_{.02}$} & {\scriptsize 88.1$_{.00}$} & {\scriptsize 88.8$_{.02}$} & {\scriptsize 86.9$_{.00}$} \\ 
\hline
\end{tabularx}
    \centering
        \medskip
    \caption{Results of different network architectures (Subscript $D$ indicates DeepLabV3+)}
    \label{table:arch}
\end{table}

\begin{figure}[t]
\begin{center}
\includegraphics[width=1\linewidth]{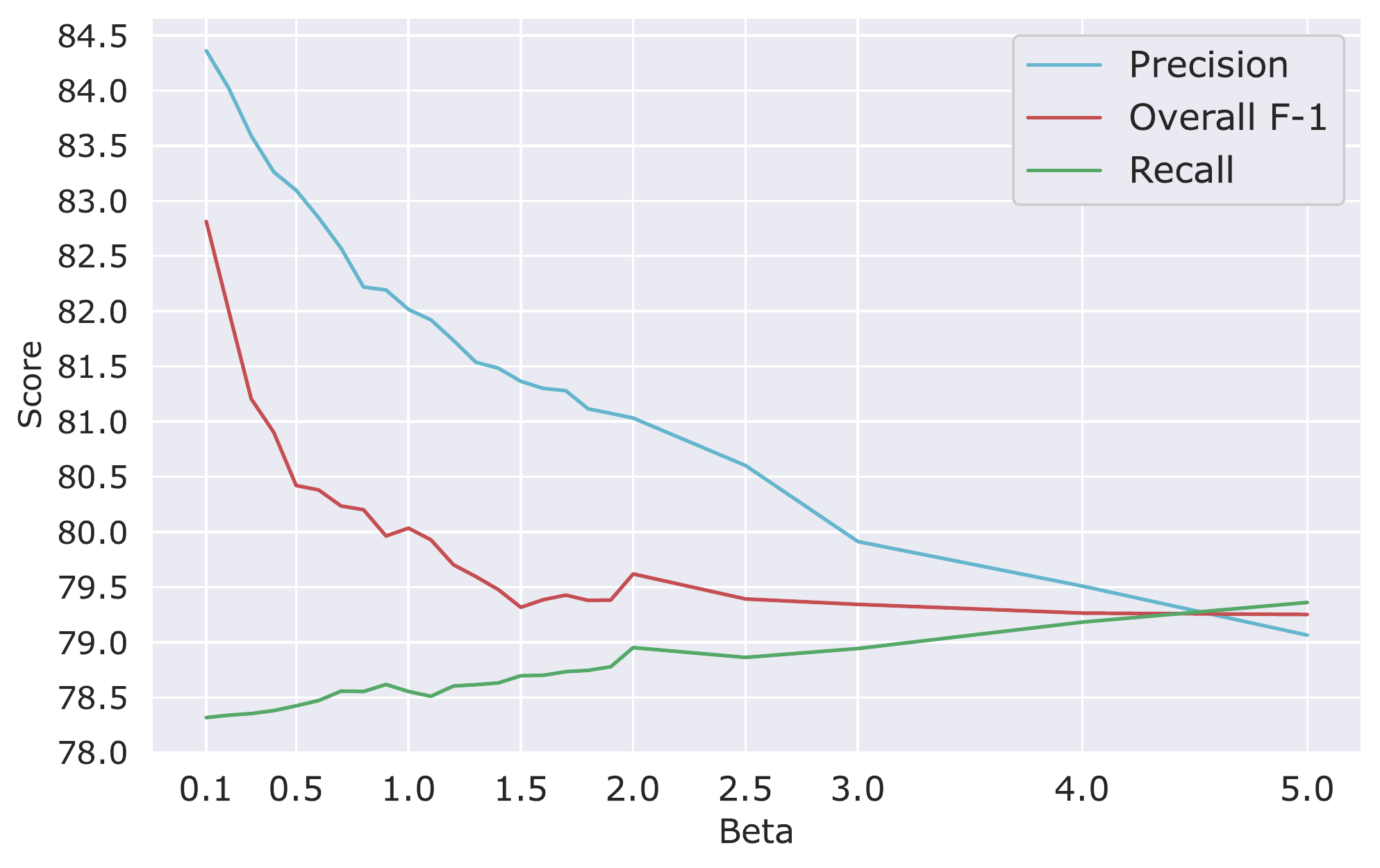}
\end{center}
  \caption{Avg. accuracy performance of three runs on different values of $\beta$ on Urban3D}
\label{fig:fbetacurve}
\end{figure}

\subsection{Experimental Evaluation and Implementation}
\noindent\textbf{Evaluation: } To evaluate each of our experiments on the datasets mentioned above, we consider the standard accuracy metrics in the fields of both semantic segmentation and binary classification: class-mean intersection-over-union (mIOU) and the F1 Score. The F1 score, also known as the Sørenson-Dice coefficient \cite{sorenson, dice} measures the balance between precision, the ratio between true positives and all positive predictions, and recall, the ratio between true positives and all ground-truth positives. In Section \ref{section:results}, we also report the average precision and recall. The mIOU metric \cite{mIOU} is a measure of the similarity between two sets using the ratio between the intersection and union of the prediction and ground-truth.\par
\noindent\textbf{General Training Details: } The experiments concerning our proposed method are 
trained for 60 epochs using a batch size of 4 with standard Stochastic Gradient Descent with a learning rate of $1e^{-4}$ and an L2 constant of $1e^{-4}$. Empirically, we found that a power of 2 when using the exponentially weighted boundary masks performs best. We select the checkpoint epoch based on the best validation loss and mIoU score.

Because Urban3D inputs are too large to fit into memory, we employ a tiling operation with a factor of 4. Each input 2048 x 2048 sample image is split into 16 inputs of 512 x 512 pixels with the same image resolution. For \textbf{only the ablation studies}, we introduce a constant amount of bias in training with the AICrowd dataset by using only 15\% of the dataset. We introduce this bias both to speed up the training process, and to continue to analyze relative performance. To account for variance, we report the mean and standard deviation (reported in subscript) across 3 experiments for each study in Tables \ref{table:arch} - \ref{table:fbetawce}.

\noindent\textbf{Cross-Dataset Training Details: } With cross-dataset training, we augment each RGB sample to 256 x 256 pixels when combining the samples from all datasets. For Urban3D and SpaceNet, we down-sample images to 512 x 512 and then split each into 4 inputs. For AICrowd, we just resize to 256 x 256. After training on the combined set for 20 epochs, we save the model at the epoch with the best loss, and then further train on each of the three datasets individually. As an implementation note, our final method was trained using this combined set for only 5 epochs and included only 10\% of the AICrowd training set.

Experiments are conducted on GPU nodes where each consists of an 8-core Intel(R) Xeon(R) CPU @ 3.00 GHz, 64 GB of RAM, and either 4 GeForce RTX 2080 Ti GPUs or 2 Tesla V100  GPUs for training models in parallel. We use PyTorch 1.6.0 to develop our methods, re-implement previously published methods, and build the training pipeline \cite{pytorch}. 

\begin{table}
\footnotesize
\begin{tabularx}{\linewidth}{ |Y|c c c c| } 
\hline
Loss & AP & AR & F-1 & mIOU \\
\hline\hline 
{U+BCE} & 80.09 & 78.04 & 78.88 & 80.31 \\
{U+F(1.0)} & 80.99 & 78.31 & 79.16 & 80.66 \\
{\hspace*{-1.5mm} \textbf{U+BCE+F(0.1)}} & \textbf{84.31} & 77.91 & \textbf{82.79} & \textbf{83.60} \\
{\hspace*{-1.5mm}U+BCE+F(1.0)} & 82.05 & 78.55 & 79.65 & 81.00 \\
{\hspace*{-1.5mm}U+BCE+F(4.0)} & 79.50 & \textbf{79.02} & 79.25 & 80.64 \\
\hline
{S+BCE} & 90.13 & 90.44 & 90.22 & 88.69 \\
{S+F(1.0)} & 90.16 & 90.91 & 90.48 & 88.96 \\
{\hspace*{-1.5mm}\textbf{S+BCE+F(0.1)}} & \textbf{92.16} & 90.37 & \textbf{92.04} & \textbf{90.65} \\
{\hspace*{-1.5mm} S+BCE+F(1.0)} & 90.04 & 91.17 & 90.61 & 89.18 \\
{\hspace*{-1.5mm} S+BCE+F(4.0)} & 89.30 & \textbf{91.76} & 90.47 & 88.96 \\
\hline
{C+BCE} & 90.19$_{.06}$ & 88.42$_{.05}$ & 89.18$_{.08}$ & 87.42$_{.08}$ \\
{C+F(1.0)} & 90.43$_{.04}$ & 89.20$_{.05}$ & 90.20$_{.05}$ & 88.55$_{.03}$ \\
{\hspace*{-1.5mm} \textbf{C+BCE+F(0.1)}} & \textbf{91.71$_{.02}$} & 89.53$_{.03}$ & \textbf{91.55$_{.03}$} & \textbf{89.96$_{.02}$} \\
{\hspace*{-1.5mm}  C+BCE+F(1.0)} & 90.74$_{.02}$ & 90.08$_{.01}$ & 90.40$_{.04}$ & 88.64$_{.00}$ \\
{\hspace*{-1.5mm} C+BCE+F(4.0)} & 89.64$_{.04}$ & \textbf{90.79$_{.01}$} & 90.25$_{.02}$ & 88.60$_{.02}$ \\
\hline
\end{tabularx}
    \centering
    \medskip
    \caption{Ablation study of BCE and F-Beta, including both losses in isolation.}
    \label{table:fbetabce}
\end{table}

\begin{table}
\footnotesize
\begin{tabularx}{\linewidth}{ |Y|c c c c| } 
\hline
Loss & AP & AR & F-1 & mIOU \\
\hline\hline 
{U+EWC} & 76.40 & \textbf{82.27} & 78.58 & 79.97 \\
{\hspace*{-1.5mm} \textbf{U+EWC+F(0.1)}} & \textbf{83.82} & 81.10 & \textbf{83.01} & \textbf{84.22} \\
{\hspace*{-1.5mm} U+EWC+F(0.5)} & 81.49 & 81.14 & 80.75 & 81.87 \\
{\hspace*{-1.5mm} U+EWC+F(1.0)} & 80.09 & 81.72 & 80.26 & 81.40 \\
{\hspace*{-1.5mm} U+EWC+F(4.0)} & 78.98 & 81.90 & 79.68 & 80.95 \\
\hline
{S+EWC} & 88.90 & 91.51 & 90.13 & 88.55 \\
{\hspace*{-1.5mm} \textbf{S+EWC+F(0.1)}} & \textbf{92.70} & 91.70 & \textbf{92.41} & \textbf{90.94} \\
{\hspace*{-1.5mm} S+EWC+F(0.5)} & 90.41 & 91.74 & 91.02 & 89.53 \\
{\hspace*{-1.5mm} S+EWC+F(1.0)} & 89.36 & 92.19 & 90.70 & 89.37 \\
{\hspace*{-1.5mm} S+EWC+F(4.0)} & 89.25 & \textbf{92.34} & 90.52 & 88.98 \\
\hline
{C+EWC} & 89.80$_{.04}$ & 90.05$_{.07}$ & 89.85$_{.03}$ & 88.00$_{.04}$ \\
{\hspace*{-1.5mm} \textbf{C+EWC+F(0.1)}} &\textbf{92.04$_{.05}$} & 91.06$_{.08}$ & \textbf{91.67$_{.07}$} & \textbf{90.61$_{.04}$} \\
{\hspace*{-1.5mm} C+EWC+F(0.5)} & 91.02$_{.03}$ & 91.24$_{.10}$ & 91.45$_{.02}$ & 89.72$_{.01}$ \\
{\hspace*{-1.5mm} C+EWC+F(1.0)} & 89.59$_{.05}$ & 91.63$_{.03}$ & 90.50$_{.05}$ & 88.90$_{.02}$ \\
{\hspace*{-1.5mm} C+EWC+F(4.0)} & 89.38$_{.02}$ & \textbf{91.82$_{.08}$} & 90.39$_{.04}$ & 88.83$_{.06}$ \\
\hline
\end{tabularx}
    \centering
    \medskip
    \caption{Ablation study of EWC and F-Beta, including EWC in isolation.}
    \label{table:fbetawce}
\end{table}

\section{Results}
\label{section:results}

\begin{figure*}
\begin{center}
\includegraphics[width=\textwidth]{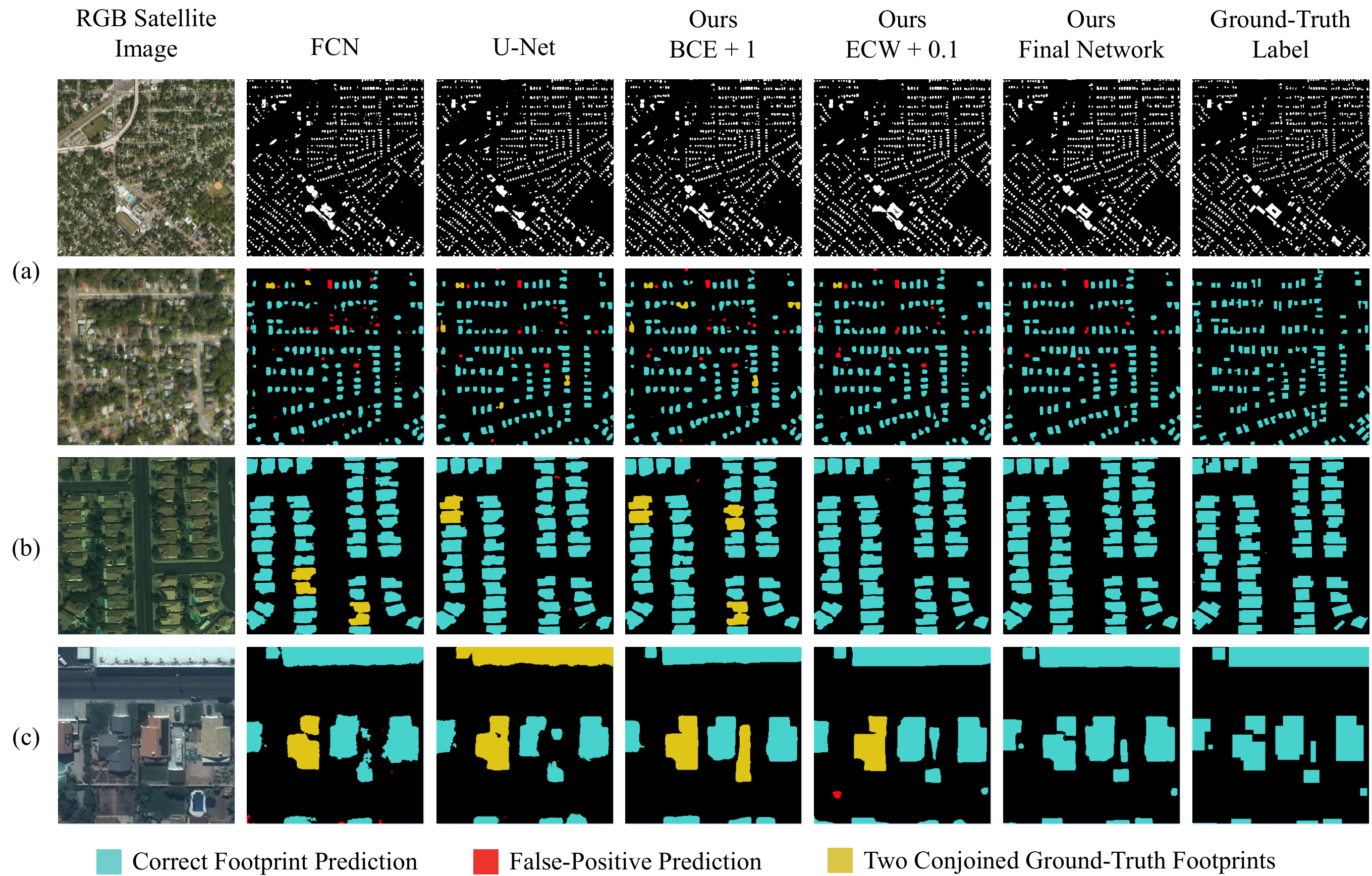}
\end{center}
  \caption{Visualizations of predictions from all datasets by four variants of our method and two baselines. \textbf{Top-Down: (a)} Urban3D [second row is a magnification of the first row for better visual comparison], \textbf{(b)} SpaceNet, \textbf{(c)} AICrowd. (Best viewed electronically)}
\label{fig:visualpredictions}
\end{figure*}

\begin{table*}[t]
\footnotesize
\begin{center}
\begin{tabularx}{0.6\textwidth}{ |Y|c|c c c c c| } 
\hline
Dataset & Method & AP & AR & F-1 & mIOU & IOU$_{y=1}$ \\
\hline\hline
\multirow{5}{4em}{Urban3D} & R2U-Net \cite{zorzi} & - & - & - & - & 69.9$_{\pm 7.1}$ \\ 
& U-Net & 84.6 & 79.2 & 81.0 & 80.8 & 65.9$_{\pm 15.1}$ \\ 
& U-Net + DEM  & \textbf{85.3} & 80.0 & 81.8 & 83.0 & 68.7$_{\pm 15.5}$ \\ 
& Ours & 83.8 & 81.1 & 83.0 & 84.2 & 70.1$_{\pm 15.2}$ \\ 
& \textbf{Ours + Cross-Dataset} & 84.5 & \textbf{81.3} & \textbf{83.4} & \textbf{84.5} & \textbf{70.2$_{\pm 15.3}$} \\ 
\hline
\multirow{4}{4em}{SpaceNet} & U-Net Ensemble \cite{cvprspacenetunetensemble} & - & - & 86.4 & - & - \\ 
& Mask-RCNN \cite{cvprspacenetmrcnn} & - & - & 88.1 & - & - \\ 
& Ours & 92.7 & 91.7 & 92.4 & 90.9 & 85.2$_{\pm 3.5}$ \\
& \textbf{Ours + Cross-Dataset} & \textbf{93.2} & \textbf{92.0} & \textbf{92.6} & \textbf{91.1} & \textbf{85.5$_{\pm 3.6}$} \\ 
\hline
\multirow{5}{4em}{AICrowd} & Mask-RCNN \cite{zorzi} & - & - & - & - & 74.2$_{\pm 14.5}$ \\
& R2U-Net \cite{zorzi} & - & - & - & - & 80.0$_{\pm 14.2}$ \\
& U-Net \cite{crowdaitds} & 94.3 & 95.4 & 94.8 & - & - \\ 
& \textbf{Ours} & 96.2 & \textbf{96.3} & \textbf{96.3} & \textbf{95.4} & \textbf{92.2$_{\pm 2.6}$} \\ 
& Ours + Cross-Dataset & \textbf{96.3} & 96.0 & 96.1 & 95.2 & 91.9$_{\pm 3.1}$ \\ 
\hline
\end{tabularx}
\end{center}
\caption{State-of-the-Art Comparisons. We use 100\% of AICrowd to train/test our final model.}
\label{table:sotac}
\end{table*}

In most tables, we use shorthand notation for datasets: \textbf{U}-Urban3D, \textbf{S}-SpaceNet, \textbf{C}-AICrowd.

\subsection{State-of-the-Art Comparisons}
We compare our method to current state-of-the-art segmentation networks by dataset. For AICrowd and Urban3D, we follow \cite{zorzi} which use an R2U-Net and Mask-RCNN followed by an extensive regularization and polygonization process. For only AICrowd, we follow the approach detailed by the winning submission that used a U-Net with a ResNet-101 encoder, in total containing 51.52M parameters \cite{crowdaitds}. Then for Urban3D, we follow the winning approach which similarly takes advantage of a U-Net architecture with a ResNet-34 encoder (24.43M parameters) but requires additional depth elevation models. We report the performance of this architecture with and without the supplemental data. For SpaceNet, we follow \cite{cvprspacenetmrcnn}, an approach similar to \cite{zorzi} that employs a Mask-RCNN (87.33M parameters) and a boundary regularizer in post-processing, and \cite{cvprspacenetunetensemble}, an approach that requires two forward passes in a U-Net Ensemble (31.53M parameters). We note that \cite{cvprspacenetunetensemble} takes advantage of multi-spectral images which offer information through 8-channels, \textbf{while our model uses only 3-channel RGB images}. Our final network contains 58.26M parameters. Table \ref{table:sotac} reports the results from all methods, and we also show the results of our final method with cross-dataset training. Additionally, we report the IOU on the building class (IOU$_{y=1}$) to compare with \cite{zorzi}. We show that despite data-level disadvantages our method performs better than all architectures, notably including the U-Net architectures across all datasets. We also note that compared to \cite{cvprspacenetmrcnn} and \cite{zorzi}, we do not require any post-processing after segmentation. Additionally, while the U-Net + DEM nears our final performance, it is at a significant advantage with an extra channel (see Figure \ref{fig:dataset}). The results we have presented in Table \ref{table:sotac} quantitatively show our proposed method is state-of-the-art, requiring a single end-to-end network and no additional channels of information.

\subsection{Network Architecture}
Table \ref{table:arch} compares our base network and the three baseline architectures.  We compare our DeepLabV3+ and Dilated ResNet network \textbf{(DRN$_D$)} \cite{deeplabv3, dilatedresnet} against three primary baselines: the DeepLabV3 + ResNet-101 \textbf{(RN$_D$)} \cite{resnet}, as it is the most common pairing in DeepLabV3+ related literature, a standard U-Net encoder-decoder as it is the base network in many related works \cite{unet}, and a vanilla fully-connected network since it is a notable architecture that has been tested on other building footprint datasets \cite{fcn, yang2018}. The results indicate that the baseline method performs quite well with precision but falls behind with recall, thus creating an imbalance that brings their F1 performance down. Although the ResNet-101 backbone does not perform as well in precision, it offers a more balanced F1 Score performance and often outperforms in mIOU.  Finally, the numerical results show that the Dilated ResNet backbone outperforms the ResNet backbone across all performance metrics. The performance of the Dilated ResNet compared to the larger ResNet-101 model supports the claims laid out in Section \ref{section:deeplab}: dilated convolutions help preserve the spatial structure of footprint predictions and expand the contextual receptive field. The results of the Dilated Res-Net backbone and baseline methods can be visualized in Figure \ref{fig:visualpredictions}. 
Empirically, the visual results show that the FCN produces noisy footprints, especially on the border, and the U-Net often produces more "under-filled" footprints, thus the lack in recall. The Dilated Res-Net backbone (labeled "BCE+1") provides significantly sharper footprints notwithstanding the conjoined predictions. 

\subsection{$\beta$ in F-Beta Measure and Exponentially Weighted Boundary Loss}
Figure \ref{fig:fbetacurve} illustrates how precision, recall, and overall F-1 score vary as we tune the $\beta$ in the F-Beta measure. Reinforcing the claims in Section \ref{section:objective}, the figure shows that recall increases with $\beta$, but at a much slower rate compared to how precision proliferates as $\beta$ decreases. From this figure, we claim that building segmentation is inherently a problem of precision, and overall performance benefits from placing more emphasis on precision than recall. We note that in addition to precision's faster rate of growth, as we bring $\beta$ closer to zero, the overall F-1 score increases exponentially. \par

We report the results of using the F-Beta Measure and the Exponentially Weighted Boundary Loss (EWC) in the form of two tables. First, Table \ref{table:fbetabce} concerns the ablation study between F-Beta and binary cross entropy (BCE). The data shows that the network does not benefit from only BCE, and the performance of using only the DICE measure ($\beta=1$) falls slightly short of the performance reached by both losses combined. Furthermore, the table illustrates the trade-off between low and high values of $\beta$ when combining the BCE and F-Beta objective functions. With high $\beta$, we achieve much better recall but suffer in all other metrics. With $\beta=0.1$, we achieve considerable gains in precision and mIOU, and despite the lack in recall we consistently achieve a higher F-1 score. \par

In Table \ref{table:fbetawce}, we report the results of F-Beta and EWC. We first note that EWC alone performs slightly worse than BCE, despite placing more emphasis on separating buildings. In addition, EWC tends to achieve higher performance in recall than precision. This is because while EWC is focused on boundaries, it is not focusing on preventing sparse, erroneous footprint predictions. On the other hand, we incur overall performance benefits when we combine EWC and F-Beta together. The relative precision and recall trade-offs of $\beta$ are still maintained when using EWC, and $\beta=0.1$ continues to have better overall F-1 and mIOU performance. Besides the increase in mIOU, when using EWC over BCE in conjunction with F-Beta we generally observe that precision tends to fall and recall improves. Consequently, this creates a more balanced performance between precision and recall, thus consistently improving the F-1 score. Figure \ref{fig:visualpredictions} shows the results of using EWC and $\beta = 0.1$. Compared to the previous iteration, we effectively resolved previously conjoined and sparse false-positive footprints across all three datasets. For the footprints that are still conjoined, we have improved separation. 

\subsection{Cross-Dataset Training}
Table \ref{table:sotac} reports the performance of our final network when using cross-dataset training. Note that despite inter-dataset variance, as illustrated by Figure \ref{fig:dataset}, the cross-dataset strategy achieved a slightly higher performance on both Urban3D and SpaceNet. The same strategy did not perform similarly on the AICrowd dataset, which leads us to believe that cross-dataset training is not as effective on datasets with a saturated presence in the combined set. Qualitatively, Figure \ref{fig:visualpredictions} shows that our final method generates footprints with a more defined shape and fewer conjoined predictions. 


\section{Conclusion}

In order to better support urban energy simulation through Deep Learning based approaches, convolutional neural networks must first be able to produce quality results with distinct footprints, and be scale invariant to the fickle nature of satellite imagery. To increase accessibility to this technology, approaches should also be robust to the lack of special wideband imagery that only few urban centers are able to produce. \par

In this paper, we present a new approach that brings Deep Learning closer to perfecting building segmentation. In the presence of class imbalance, we demonstrate that our proposed F-Beta measure is efficacious at resolving false-positives without de-prioritizing the common class. We show that combining F-Beta with our EWC loss creates a powerful tool to separate distinct entities that were previously unified. By using these advances and a DeeplabV3+ architecture with a Dilated ResNet backbone, we achieve state-of-the-art performance on three standard datasets and provide a strong network which can assist urban planners worldwide in climate related decision making.

{\small
\bibliographystyle{ieee_fullname}
\bibliography{rgbfinal}
}

\pagebreak
\begin{alphasection}

\section{Supplemental Materials}

\subsection{Additional Visualizations}

Besides Figure \ref{fig:visualpredictions}, we provide additional visualizations of model predictions on the Urban3D \cite{urban3d}, SpaceNet \cite{spacenet}, and AICrowd \cite{crowdAI} datasets in Figure \ref{fig:visualpredictionspt2}.

\subsection{Further Analysis}

As discussed in Section {\color{red} 4} of the main paper and supported by Figures \ref{fig:visualpredictions} and \ref{fig:visualpredictionspt2}, the predictions by the fully-convolutional network \cite{fcn} frequently exhibit noisy and sparse predictions. Although the U-Net \cite{unet} resolves almost all of these noisy predictions, thus the high average precision, the model does not effectively capture the geometry of footprints which brings down the average recall. This is especially evident in Figure \ref{fig:visualpredictionspt2} (b)(i), where the U-Net is unable to capture the smaller buildings in the upper portion of the center neighborhood in the satellite image. 

The first iteration of our network (column "BCE + 1") solves the issues the U-Net gave rise to as it is now able to capture more of the building footprints. However, as discussed in the main paper, this iteration often creates conjoined footprint predictions and curved building geometries. The next iteration of our network (column "ECW + 0.5") employs the F-Beta measure and the exponentially weighted boundary loss to encourage the network to separate distinct footprints and create sharper edges. These claims are observable primarily in the predictions of Urban3D and SpaceNet examples (Figure \ref{fig:visualpredictionspt2} (a) and (b)). Notice that by using a new objective, we were able to separate most of the combined footprints from the previous iteration and create straighter edges. Finally, the last iteration of our network (column "Final Network") involved using a cross-task training strategy on only the Urban3D and SpaceNet datasets to resolve any remaining prediction artifacts. Observe from Figure \ref{fig:visualpredictionspt2} that this strategy consistently produces predictions that improves upon the previous iteration by separating footprints that were still combined, enforces sharper corners, and cleans up most remaining false positives.

\begin{figure*}[h!]
\begin{center}
\includegraphics[width=\textwidth]{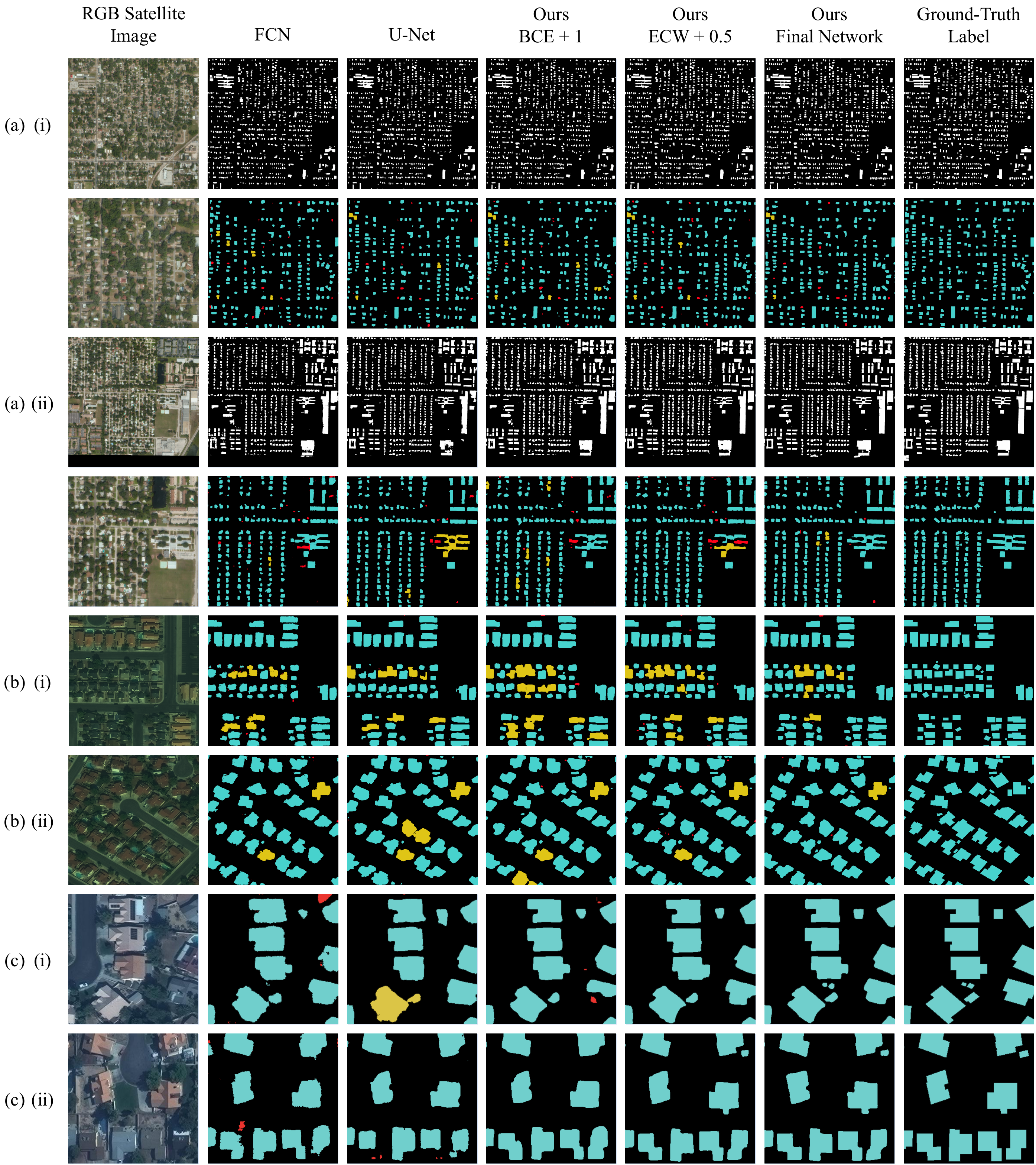}
\end{center}
  \caption{Additional \textbf{two} visualizations of predictions from each of the three datasets used in this paper. \textbf{Top-Down: (a)} Urban3D [the second row of each prediction is a magnification of the first row for better visual comparison], \textbf{(b)} SpaceNet, \textbf{(c)} AICrowd. Turquiouse buildings are correctly predicted footprints, yellow predictions are two conjoined ground-truth footprints, and red predictions are false positive footprints.}
\label{fig:visualpredictionspt2}
\end{figure*}

\end{alphasection}

\end{document}